\documentclass[letterpaper]{article} 
\usepackage{aaai24}  
\usepackage{times}  
\usepackage{helvet}  
\usepackage{courier}  
\usepackage[hyphens]{url}  
\usepackage{graphicx} 
\usepackage{natbib}  
\usepackage{caption} 
\usepackage{xcolor}

\usepackage{algorithm}
\usepackage{algorithmic}
\usepackage{booktabs}
\usepackage{subfig}
\usepackage{multirow}
\usepackage{amsthm}
\usepackage{bm}
\usepackage{color}
\usepackage{colortbl}
\usepackage{makecell}
\usepackage{amsmath}

\usepackage{newfloat}
\usepackage{listings}
\DeclareCaptionStyle{ruled}{labelfont=normalfont,labelsep=colon,strut=off} 
\floatstyle{ruled}
\newfloat{listing}{tb}{lst}{}
\floatname{listing}{Listing}
\nocopyright

\title{Enhancing  Infrared Small Target Detection Robustness with Bi-Level Adversarial Framework}
\author{
    Zhu Liu,
   Zihang Chen,
    Jinyuan Liu,
    Long Ma,
    Xin Fan, 
    Risheng Liu\textsuperscript{\rm 1}
}
\affiliations{
    \textsuperscript{\rm 1}Dalian University of Technology\\
    liuzhu@mail.dlut.edu.cn, chenzi$\_$hang@mail.dlut.edu.cn, atlantis918@hotmail.com, rsliu@dlut.edu.cn

}

\usepackage{bibentry}

\begin{document}

\maketitle

\begin{abstract}
	The detection of small infrared targets against blurred and cluttered backgrounds has remained an enduring challenge. In recent years, learning-based schemes have become the mainstream methodology to establish the mapping directly. However, these methods are susceptible to the inherent complexities of changing backgrounds and real-world disturbances, leading to unreliable and compromised target estimations. In this work, we propose a bi-level adversarial framework to promote the robustness of detection in the presence of distinct corruptions. We first propose a bi-level optimization formulation to introduce dynamic adversarial learning. Specifically, it is composited by the learnable generation of corruptions to maximize the losses as the lower-level objective and the robustness promotion of detectors as the upper-level one. We also provide a hierarchical reinforced learning strategy to discover the most detrimental corruptions and balance the performance between robustness and accuracy. To better disentangle the corruptions from salient features, we also propose a spatial-frequency interaction network for target detection. Extensive experiments demonstrate our scheme remarkably improves 21.96\% IOU across a wide array of corruptions and notably promotes 4.97\% IOU on the general benchmark. The source codes are available at \url{https://github.com/LiuZhu-CV/BALISTD}.
\end{abstract}

\section{Introduction}
Infrared Small Target Detection (ISTD) has been a vital component for infrared search and tracking systems, which refers to discovering tiny targets with low contrast from complex infrared backgrounds. This technique has attracted widespread attention and been widely leveraged for diverse real-world applications, such as military surveillance~\cite{zhao2022single,sun2022drone}, 
traffic monitoring~\cite{zhang2022rkformer,liu2021learning} and marine rescue~\cite{zhang2022exploring,liu2020real}.

Different from general object detection, as a long-standing and challenging task, ISTD is limited by the nature characteristics of infrared imaging with small targets.  (1) Complex background with corruptions:  motion blur and high noise often occur because of the fast-moving and severe environments. Furthermore, due to the different Image Signal Processing (ISP) of diverse devices, thermal information and contrast  are totally different.  (2) Small, dim, shapeless, and texture-less of infrared targets: they are usually with 
low contrast and low Signal-to-Clutter Ratio (SCR), resulting in submergence under complex background.

\begin{figure}[t]
	\centering
	\begin{tabular}{c@{\extracolsep{0.1em}}c} 
		{\includegraphics[width=0.47\textwidth]{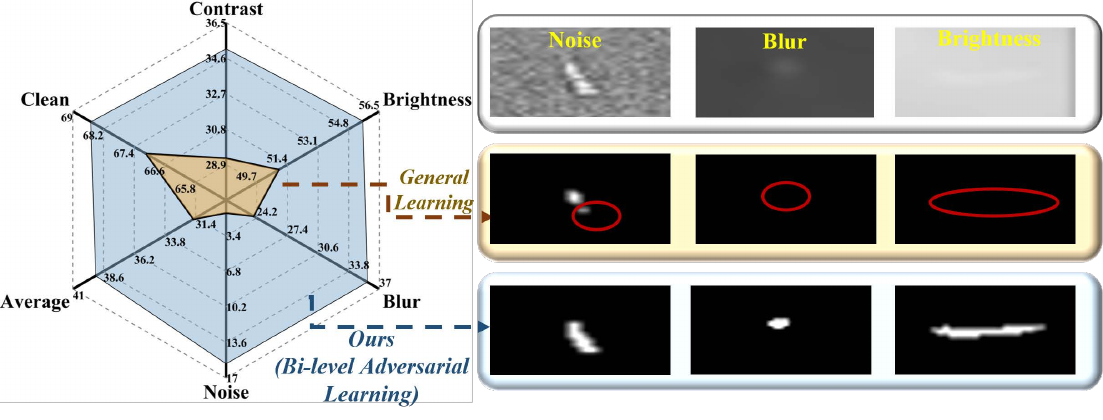}}\\
	\end{tabular}
	\vspace{-0.3cm}
	\caption{Illustration of our core contributions to address diverse corruptions for infrared small target detection by numerical accuracy (IOU) and visual comparisons. Compared with general end-to-end learning, the proposed bi-level adversarial training realizes the remarkable promotion.}
	\label{fig:fristfig}
	
\end{figure}

In the past decades, numerous efforts have been devoted to this task, which can be roughly divided into two categories, \textit{i.e.,} conventional numerical methods and learning-based methods. For instance, low-rank representation~\cite{gao2013infrared,zhang2018infrared,zhang2019infrared,liu2012fixed,wu2019essential}, local contrast-based~\cite{moradi2018false}, and filtering~\cite{qin2019infrared} are three typical kinds of schemes based on diverse domain knowledge and handcrafted priors. Nevertheless, these methods heavily rely on the manual adjustment of hyper-parameters and dedicated feature extraction based on expert engineering. The complex numerical iterations and feature construction limit their performances in real-world detection scenarios.

Since the effective feature extraction and data fitting, learning-based schemes realize remarkable improvements with many effective architectures~\cite{liu2021retinex,dai2021asymmetric,piao2019depth,li2022dense,piao2020a2dele,zhang2020select} in recent years. For instance,  asymmetric contextual modulation~\cite{dai2021asymmetric} is introduced to investigate the semantic features and spatial texture details. In order to balance the miss detection and false alarm, generative adversarial network~\cite{wang2019miss} is leveraged for ISTD.
Introducing the shape information of small targets,~\cite{zhang2022isnet} proposed an effective shape-aware network. Transformer  is also utilized for ISTD task based on Runge-Kutta approximation~\cite{zhang2022rkformer}.  Lastly,~\cite{ying2023mapping} designed a flexible learning strategy with label evolution to efficiently reduce the annotations.


However,  we argue that two exist two major stumbling stones that hinder the development of learning-based methods.
Firstly, as for the training strategies, we emphasize that there lack of effective mechanisms to address these 
factors of corruption~\cite {ying2023mapping}. Most existing methods are based on end-to-end learning with large labeled datasets. However, these methods are vulnerable to
changes in data distributions and corruptions, leading to misdetection and weak generalization.
Secondly, as for the architecture design of learning-based methods, there is a lack of robust architectures to distinguish the salient representations from corrupted and cluttered features.  The proposed mechanisms are only focused on accuracy promotion and are at risk of being vulnerable to corruption. The degraded perturbations cannot be removed utilizing the currently proposed modules. From these observations, our goal is to propose a general robust framework to promote robustness and generalization both from the training strategy and architectural perspectives.

To partially alleviate these issues, we propose a bi-level adversarial framework to automatically discover the sample-correlated corruptions for the robustness promotion of ISTD. We make the first attempts to incorporate the influence of various corruptions into the optimization of  ISTD tasks, as shown in Figure.\ref{fig:fristfig}. In detail, we devise a bi-level optimization framework with two adversarial objectives. First, the former goal is the generation of sample-related corruptions, which aims to fool the detection network to estimate inaccurate results. Second, the later process aims to balance the robustness and accuracy based on the corrupted and clean samples. The adversarial principle lies in maximizing training losses by corruption generation and minimizing losses by detection optimization. In order to solve this adversarial procedure, we present a  hierarchical reinforced strategy to approximately
optimize these goals, dividing into two training procedures including the sample generation based on the evaluations of detection robustness and the trade-off learning between robustness and accuracy. Then we propose a spatial-frequency interaction network to disentangle harmful components of degradation both in the spatial and Fourier domain.
Our contributions are summarized as follows:

\begin{itemize}
	\item  By formulating the corruption generation and model robustness as two adversarial goals, we propose a bi-level adversarial framework. To the best of our knowledge, it is the first attempt to systematically investigate the robustness of ISTD models under various corruptions. 
	\item From the training side, we propose a hierarchical reinforced strategy to guide the optimization of corruption strategy generation and detection training, which involves the balance between the robustness of  ISTD model and the task accuracy to solve the  optimization.
	\item From the architectural side, we propose a spatial-frequency interaction network to separate the degradation from discriminative features, which can effectively promote robustness under diverse corruptions.
	\item Comprehensive experiments show that the proposed scheme empirically not only achieves consistent robustness under corruption but also drastically improves performance on three general benchmarks. As a plug-and-play framework, our paradigm also can strengthen the performance of other current advanced models.
	
\end{itemize}
\begin{figure*}[htb]
	\centering
	\setlength{\tabcolsep}{1pt} 
	
	\includegraphics[width=0.99\textwidth,]{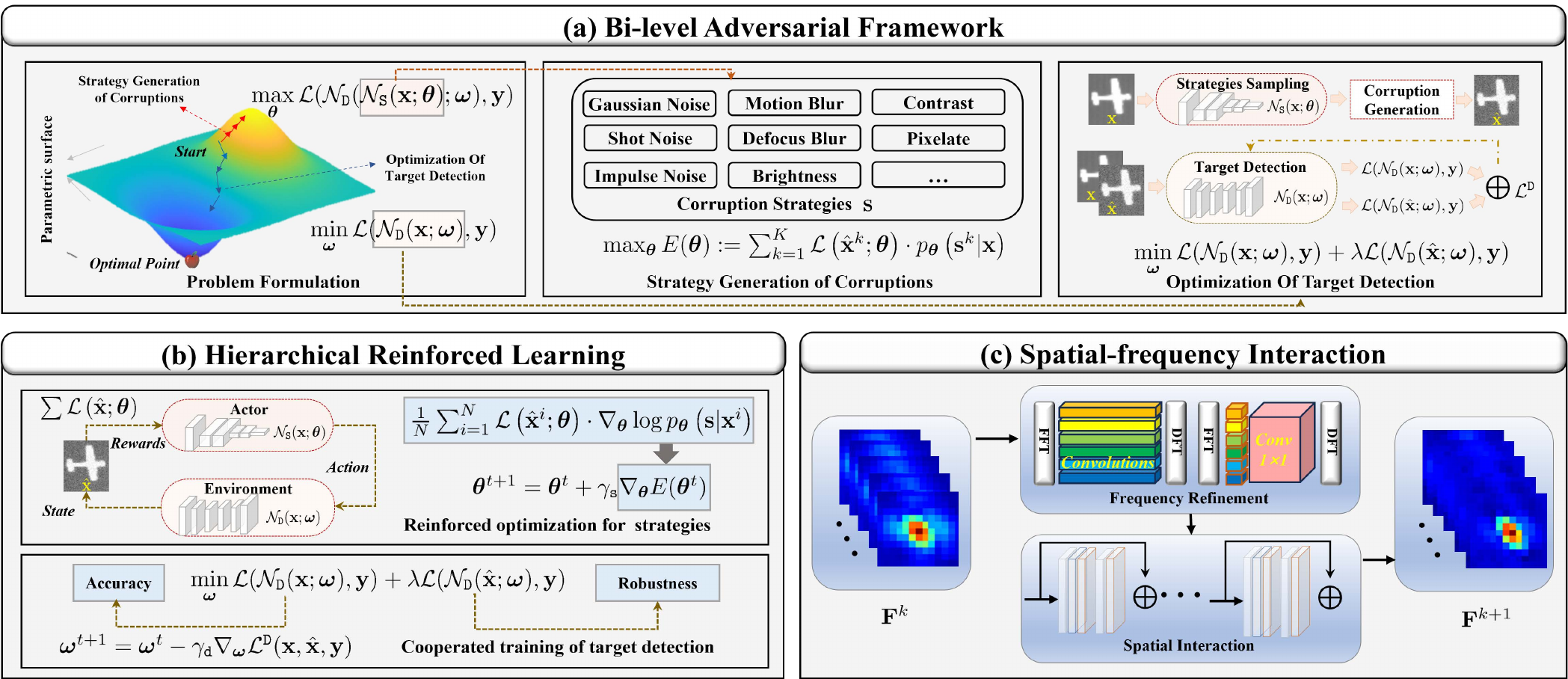}
	
	\caption{Schematic graph of the proposed bi-level adversarial framework. We first illustrate the bi-level formulation including the strategy generation of corruptions and optimization of the ISTD network in subfigure (a). In subfigure (b), we present the hierarchical reinforced learning for strategy generation and small target detection.  Lastly, we depict the concrete networks of spatial-frequency interaction in subfigure (c).}
	
	\label{fig:workflow}
\end{figure*}
%
%

\section{Proposed Method}
In this section, we  elaborate the definition of bi-level adversarial formulation and basic pipeline. Then we present the hierarchical reinforced learning for  strategy for the optimization  and architectures of proposed frameworks including strategy generation and target detection.

\subsection{Bi-level Adversarial Framework}
Existing learning-based methods seldom consider the solutions to improve robustness under corruption, and design specialized networks to directly learn the latent data correspondences. 
Considering the defence of corruptions as one adversarial game,  we propose a bi-level  formulation~\cite{liu2021investigating,ma2023bilevel,liu2023bilevel,liu2023multi} to automatically generate the specific corruptions for samples from a learnable perspective instead of utilizing these manually designed augmentations. In detail, we introduce two competitive networks, including the strategy generation network $\mathcal{N}_\mathtt{S}$ with parameters $\bm{\theta}$ and detection network $\mathcal{N}_\mathtt{D}$ with parameters $\bm{\omega}$ to conduct the adversarial procedures. The former $\mathcal{N}_\mathtt{S}$ provides one corruption strategy to attack the detection for unreliable estimation.
Specifically, we denote $\mathbf{s}$ as the corruption strategy, which can be defined as $\mathbf{s}:=\{s_{1},\cdots s_{n}\}$. $s_{i}$ represents one parameter for this category of corruption. Given one image $\mathbf{x}$ and network parameters $\bm{\theta}$, we can obtain one sample-independent corruption $\mathbf{s}$ from the conditional distribution $p\left(\mathbf{s} | \mathbf{x};\bm{\theta}\right)$.
Meanwhile, $\mathcal{N}_\mathtt{D}$ leverages the selected corruptions to improve the robustness.

Letting the above intuition precise, we can formulate the optimization of both competitors as:
\begin{align}
&	\min\limits_{\bm{\omega}}  \mathcal{L}(\mathcal{N}_\mathtt{D}(\mathbf{x};\bm{\omega}),\mathbf{y}) + {\lambda} \mathcal{L}(\mathcal{N}_\mathtt{D}(\hat{\mathbf{x}};\bm{\omega}),\mathbf{y}),\label{eq:main}\\
&	\mbox{ s.t. } \left\{
\begin{aligned}
\hat{\mathbf{x}} &= \mathcal{N}_\mathtt{S}(\mathbf{x};\bm{\theta}^{*}), \\
\bm{\theta}^{*} &= \arg\max\limits_{\bm{\theta}} \mathcal{L}(\mathcal{N}_\mathtt{D}(\mathcal{N}_\mathtt{S}(\mathbf{x};\bm{\theta});\bm{\omega}),\mathbf{y}),
\end{aligned}
\right.	
\label{eq:constraint}
\end{align}
where  $\mathcal{L}$ is the ISTD-related losses and $\lambda$ denotes the trade-off parameter. $\mathbf{x}$, $\hat{\mathbf{x}}$, and $\mathbf{y}$ are the clean, corrupted samples and labels, respectively. We utilize the upper-level objective (\textit{i.e.,} Eq.~\eqref{eq:main}) to balance the robustness and detection accuracy of $\mathcal{N}_\mathtt{D}$. Moreover, we introduce the nested constraint (\textit{i.e.,} Eq.~\eqref{eq:constraint}) by the automatic selection of specialized corruptions based on the strategy generation from $\mathcal{N}_\mathtt{S}$.

We argue that bi-level adversarial learning has a significant impact on the robustness promotions of ISTD by a dynamic competitive game. In the initial stages of optimization, the accuracy of the detection network is susceptible to a small degree of corruption. The generation network can easily select effective strategies (either weak or strong degrees) to realize the goal of  Eq.~\eqref{eq:constraint}. As the training progresses, the $\mathcal{N}_\mathtt{D}$ can be more robust by the optimization of Eq.~\eqref{eq:main}. The strategy generation must produce stronger strategies to adapt the ISTD network. This dynamic game can contribute to a gradual promotion of the robustness of $\mathcal{N}_\mathtt{D}$, which is more flexible and effective compared with handcrafted ones.

\subsection{Hierarchical Reinforced Learning}
There are two limitations to solve the above optimization. Firstly, the exact solutions are huge computations and complexities~\cite{liu2021value,liu2020bilevel,liu2023task}. Recent min-max optimization (\textit{e.g.,} Generative Adversarial Network (GAN)~\cite{goodfellow2014generative,liu2022target} and Adversarial Training (AT)~\cite{zhang2022revisiting,jia2022adversarial})  always leverage the alternative learning strategies to approximately solve these objectives.
Secondly,  another 
stumbling block is to obtain the gradient of the strategy network.
The procedure of generation is not differentiable, including some non-differentiable parameters such as the degrees and categories of operations. 

Thus, based on the alternative optimization, we propose a hierarchical reinforced learning scheme to address this competitive formulation (\textit{i.e.,} Eq.~\eqref{eq:main} and Eq.~\eqref{eq:constraint}), which can be divided into two parts, the reinforced optimization for strategies and cooperated learning for detection.


\noindent \textbf{Reinforced optimization for  strategies.}  Considering $K$ corruptions, we  reformulate the sub-problem (Eq.\eqref{eq:constraint}) as 
\begin{equation}\label{eq:sub}
\max_{\bm{\theta}} E(\bm{\theta}) : =   \sum_{k=1}^K \mathcal{L}\left(\hat{\mathbf{x}}^k; \bm{\theta}\right) \cdot  p_{\bm{\theta}}\left(\mathbf{s}^k | \mathbf{x}\right),
\end{equation}
where from the viewpoint of reinforced learning, $\mathcal{L}$ represents the rewards and $E(\bm{\theta})$ is the summation of expectation, given $\bm{\theta}$ and $\bm{\omega}$. The strategy generation network plays the role of actor to generate corresponding actions (strategies) facing with the changes of rewards.

Based on this observation, we can introduce policy gradient algorithm~\cite{bai2023achieving} to compute the derivative, which can be  written as follows:
\begin{align}
\nabla_{\bm{\theta}} E(\bm{\theta}) & = \sum_{k=1}^K \mathcal{L}\left(\hat{\mathbf{x}}^k ; \bm{\theta}\right) \cdot  \nabla_{\bm{\theta}} p_{\bm{\theta}}\left(\mathbf{s}^k | \mathbf{x}\right)\\
& = \sum_{k=1}^K \mathcal{L}\left(\hat{\mathbf{x}}^k ; \bm{\theta}\right) p_{\bm{\theta}}\left(\mathbf{s}^k | \mathbf{x}\right) \nabla_{\bm{\theta}} \log p_{\bm{\theta}}\left(\mathbf{s}^k | \mathbf{x}\right)
\end{align}
\begin{figure*}[htb]
	\centering
	\setlength{\tabcolsep}{1pt}
	\begin{tabular}{cccccccc}
		\includegraphics[width=0.12\textwidth,height=0.085\textheight]{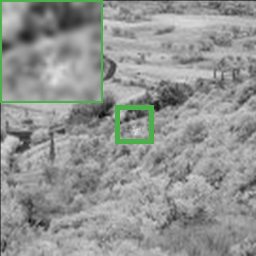}
		&\includegraphics[width=0.12\textwidth,height=0.085\textheight]{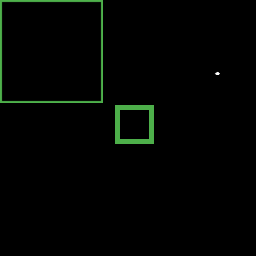}
		&\includegraphics[width=0.12\textwidth,height=0.085\textheight]{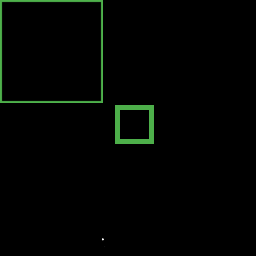}
		&\includegraphics[width=0.12\textwidth,height=0.085\textheight]{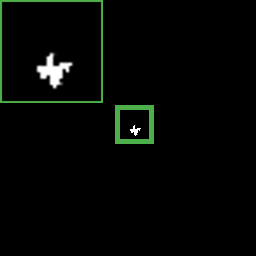}
		&\includegraphics[width=0.12\textwidth,height=0.085\textheight]{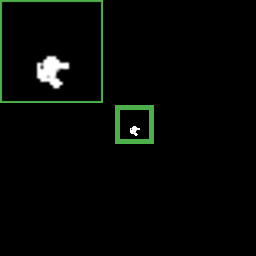}
		&\includegraphics[width=0.12\textwidth,height=0.085\textheight]{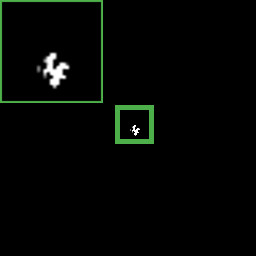}
		&\includegraphics[width=0.12\textwidth,height=0.085\textheight]{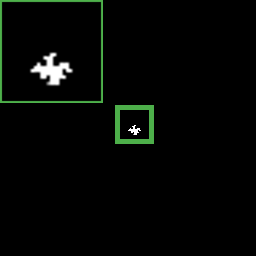}
		&\includegraphics[width=0.12\textwidth,height=0.085\textheight]{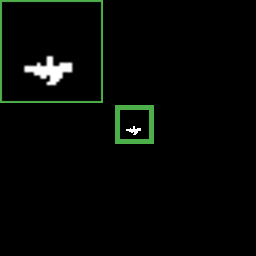}\\
		
		
		\includegraphics[width=0.12\textwidth,height=0.085\textheight]{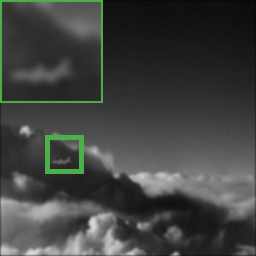}
		&\includegraphics[width=0.12\textwidth,height=0.085\textheight]{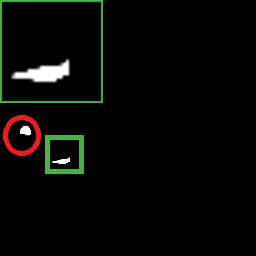}
		&\includegraphics[width=0.12\textwidth,height=0.085\textheight]{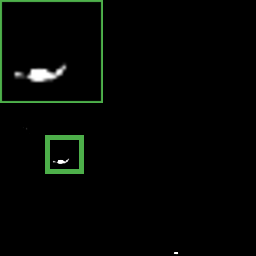}
		&\includegraphics[width=0.12\textwidth,height=0.085\textheight]{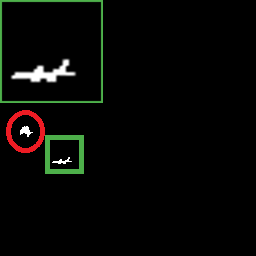}
		&\includegraphics[width=0.12\textwidth,height=0.085\textheight]{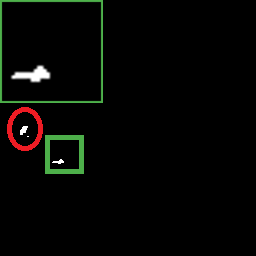}
		&\includegraphics[width=0.12\textwidth,height=0.085\textheight]{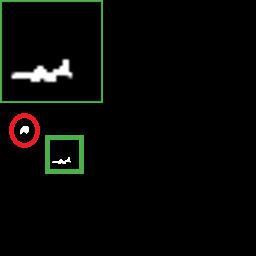}
		&\includegraphics[width=0.12\textwidth,height=0.085\textheight]{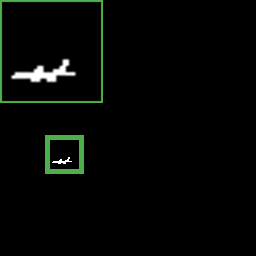}
		&\includegraphics[width=0.12\textwidth,height=0.085\textheight]{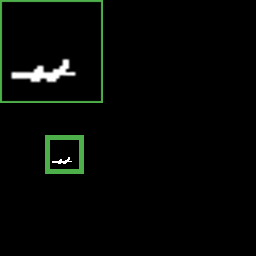}\\

		\includegraphics[width=0.12\textwidth,height=0.085\textheight]{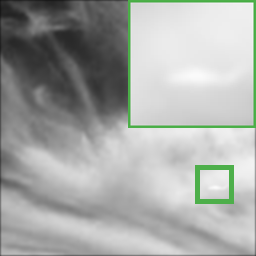}
		&\includegraphics[width=0.12\textwidth,height=0.085\textheight]{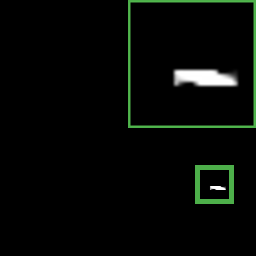}
		&\includegraphics[width=0.12\textwidth,height=0.085\textheight]{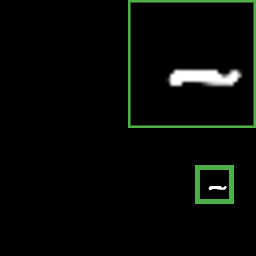}
		&\includegraphics[width=0.12\textwidth,height=0.085\textheight]{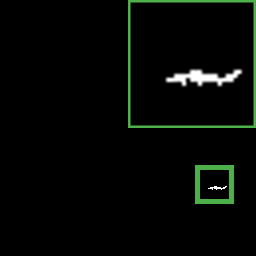}
		&\includegraphics[width=0.12\textwidth,height=0.085\textheight]{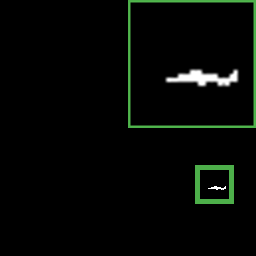}
		&\includegraphics[width=0.12\textwidth,height=0.085\textheight]{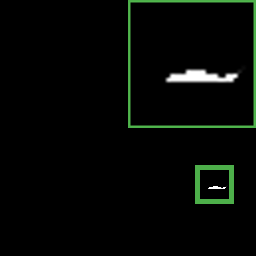}
		&\includegraphics[width=0.12\textwidth,height=0.085\textheight]{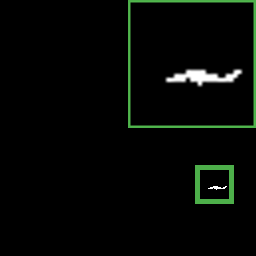}
		&\includegraphics[width=0.12\textwidth,height=0.085\textheight]{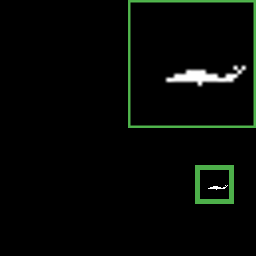}\\
		
		\footnotesize	Infrared image&\footnotesize ACM &\footnotesize ALCNet &\footnotesize DNANet &\footnotesize ISNet &\footnotesize RDIAN&\footnotesize Ours$^{*}$&\footnotesize Ground Truth
		\\
	\end{tabular}
	
	\caption{Visual comparison of different ISTD approaches on three challenging scenarios.}
	\label{fig:compare}
\end{figure*}
The approximated computation can be written as:
\begin{equation}
\nabla_{\bm{\theta}} E(\bm{\theta}) \approx \frac{1}{N} \sum_{i=1}^N \mathcal{L}\left(\hat{\mathbf{x}}^i; \bm{\theta}\right) \cdot \nabla_{\bm{\theta}} \log p_{\bm{\theta}}\left(\mathbf{s} | \mathbf{x}^i\right),
\end{equation}
where $N$ denotes the  number of sampling in one batch.
In order to maximize the objective Eq.~\eqref{eq:sub}, we utilize the gradient ascent to update the corresponding parameters, \textit{i.e.,}
\begin{equation}\label{eq:strategy}
\bm{\theta}^{t+1} = \bm{\theta}^{t} + \gamma_\mathtt{s}\nabla_{\bm{\theta}} E(\bm{\theta}^{t}),
\end{equation}
where $\gamma_\mathtt{s}$ denotes the learning rate of strategy network.
\begin{algorithm}[htb] 
	\caption{Hierarchical Reinforced Learning (HRL).}\label{alg:framework}
	\begin{algorithmic}[1] 
		\REQUIRE Clean datasets with $\{\mathbf{x},\mathbf{y}\}$,  and other necessary hyper-parameters.
		
		\WHILE {not converged}
		
		\STATE \% \emph{Cooperated training of target detection.} 
		\STATE Generating the corrupted samples by $\mathcal{N}_\mathtt{S}$ and setting batch as 
		$\{\mathbf{x}_{1}, \mathbf{y}_{1}, \cdots \hat{\mathbf{x}}_{N}, \mathbf{y}_{N} \}$;
		
		\STATE $\bm{\omega}^{t+1} = \bm{\omega}^{t} -\gamma_\mathtt{d}\nabla_{\bm{\omega}} \mathcal{L}^\mathtt{D}(\mathbf{x},\hat{\mathbf{x}},\mathbf{y})$;
		
		\STATE \% \emph{Reinforced learning of strategy generation.}
		\STATE $\bm{\theta}^{t+1} = \bm{\theta}^{t} + \gamma_\mathtt{s}\nabla_{\bm{\theta}} E(\bm{\theta}^{t})$ by policy gradient;
		\ENDWHILE
		
		\RETURN  $\bm{\omega}^{*}$.
	\end{algorithmic}
\end{algorithm}

\noindent \textbf{Cooperated training for target detection.} The main objective is to improve
the robustness of target detection. We utilize the standard adversarial training based on the trade-off balance between clean and degraded samples. Denoted the whole objective as $\mathcal{L}^\mathtt{D}$, the update of detection network can be formulated as
\begin{equation}\label{eq:detection}
\bm{\omega}^{t+1} = \bm{\omega}^{t} -\gamma_\mathtt{d}\nabla_{\bm{\omega}} \mathcal{L}^\mathtt{D}(\mathbf{x},\hat{\mathbf{x}},\mathbf{y}),
\end{equation}
where $\gamma_\mathtt{d}$ denotes the learning rate of detection network. The whole procedure is summarized in Alg.~\ref{alg:framework}.

\subsection{Architectures of Proposed Framework}

\noindent \textbf{Strategy generation network.} This learnable module is to generate sample-related corrupted strategies (\textit{i.e.,} predicting the probability of strategies). We adopt the general classifier structure. It consists of five convolution blocks and one layer of fully-connected unit. 

\noindent \textbf{Spatial-frequency interaction.} Existing schemes for ISTD mostly design mechanisms to extract salient features, ignoring the disentanglement between corrupted and distinguishable representations, which are ineffective to distinguish the targets from the  low contrast and  SCR of backgrounds. 

Recently, there is some literature~\cite{liu2021searching,zhou2022deep,zhou2023fourmer,liu2023paif} investigating  the frequency space based on  Fourier transform, which is capable of separating the degradation and adversarial artifacts from the clean features, because of the global modeling property. Inspired by these observations, we design a flexible Spatial-Frequency Interaction Module (SFIM), which can be easily embedded into existing methods to improve performance.

As shown in Fig.~\ref{fig:workflow} (c),  given one feature $\mathbf{F}^{k}$, we first utilize  Fourier transform to obtain the real and imaginary components, \textit{i.e.,} $\mathbf{F}^{k}_{I},\mathbf{F}^{k}_{R} = \mathcal{F}(\mathbf{F}^{k})$, then we leverage the parallel structure with cascaded convolutions to refine the features in the frequency domain. In detail, we first utilize the spatial convolutions for each channel of frequency features to model the spatial correlation. Then we leverage the $1\times 1$ convolution
to investigate the channel relations of frequency. 
Lastly, we perform the inverse DFT to recover the feature into the spatial domain: $\mathbf{F}^{k+1}  = \mathcal{F}^{-1}(\mathbf{F}^{k}_{I},\mathbf{F}^{k}_{R})$.
After that, we apply the spatial interaction by one residual block to gradually optimize features. Considering the DNANet~\cite{li2022dense} as the baseline, we replace the spatial-channel attention modules of the baseline with SFIM.

\begin{table*}[htb]
	\centering
	\footnotesize
	\renewcommand{\arraystretch}{1.1}
	\setlength{\tabcolsep}{2.5mm}{
		\begin{tabular}{|c|ccc|ccc|ccc|ccc|}
			\hline
			\multirow{2}{*}{Methods} & \multicolumn{3}{c|}{NUAA}                        & \multicolumn{3}{c|}{NUDT}                            & \multicolumn{3}{c|}{IRSTD-1K}   & \multicolumn{3}{c|}{Average}                           \\ \cline{2-13} 
			
			& \multicolumn{1}{c|}{\cellcolor{gray!20}IOU$\uparrow$} & \multicolumn{1}{c|}{\cellcolor{gray!20}P$_{d}$$\uparrow$} & \multicolumn{1}{c|}{\cellcolor{gray!20}F$_{a}$$\downarrow$} & \multicolumn{1}{c|}{\cellcolor{gray!20}IOU$\uparrow$} & \multicolumn{1}{c|}{\cellcolor{gray!20}P$_{d}$$\uparrow$} & \multicolumn{1}{c|}{\cellcolor{gray!20}F$_{a}$$\downarrow$}& \multicolumn{1}{c|}{\cellcolor{gray!20}IOU$\uparrow$} & \multicolumn{1}{c|}{\cellcolor{gray!20}P$_{d}$$\uparrow$} & \multicolumn{1}{c|}{\cellcolor{gray!20}F$_{a}$$\downarrow$} & \multicolumn{1}{c|}{\cellcolor{gray!20}IOU$\uparrow$} & \multicolumn{1}{c|}{\cellcolor{gray!20}P$_{d}$$\uparrow$} & \multicolumn{1}{c|}{\cellcolor{gray!20}F$_{a}$$\downarrow$} \\ \hline
			Top-Hat                        & \multicolumn{1}{c|}{7.14} & \multicolumn{1}{c|}{79.84} & 1012.00 & \multicolumn{1}{c|}{20.72} & \multicolumn{1}{c|}{78.41} & 166.70 & \multicolumn{1}{c|}{10.06} & \multicolumn{1}{c|}{75.11} & 1432.00  & \multicolumn{1}{c|}{12.64} & \multicolumn{1}{c|}{77.79} & 870.23 \\ \hline
			MSPCM   & \multicolumn{1}{c|}{12.38} & \multicolumn{1}{c|}{83.27} & 17.77 & \multicolumn{1}{c|}{5.86} & \multicolumn{1}{c|}{55.87} & 115.96 & \multicolumn{1}{c|}{7.33} & \multicolumn{1}{c|}{60.27} & 15.24  & \multicolumn{1}{c|}{7.23} & \multicolumn{1}{c|}{61.53} & 49.66 \\ \hline
			%
			NRAM   & \multicolumn{1}{c|}{12.35} & \multicolumn{1}{c|}{75.67} & {\textbf{7.89}} & \multicolumn{1}{c|}{7.42} & \multicolumn{1}{c|}{58.31}
			& 15.28& \multicolumn{1}{c|}{4.24} & \multicolumn{1}{c|}{49.16} & {\textbf{5.58}}
			& \multicolumn{1}{c|}{8.00} & \multicolumn{1}{c|}{64.38} & {\textbf{9.58}} \\ \hline
			IPI   & \multicolumn{1}{c|}{30.58} & \multicolumn{1}{c|}{87.45} & 25.38 & \multicolumn{1}{c|}{23.24} & \multicolumn{1}{c|}{79.15} & 80.87 & \multicolumn{1}{c|}{12.77} & \multicolumn{1}{c|}{69.02} & 173.39
			& \multicolumn{1}{c|}{22.20} & \multicolumn{1}{c|}{78.54} & 93.21 \\ \hline
			
			PSTNN   & \multicolumn{1}{c|}{23.02} & \multicolumn{1}{c|}{77.95} & 27.44 & \multicolumn{1}{c|}{14.87} & \multicolumn{1}{c|}{66.98} & 43.78 & \multicolumn{1}{c|}{9.94} & \multicolumn{1}{c|}{55.56} & 23.48 
			& \multicolumn{1}{c|}{15.94} & \multicolumn{1}{c|}{66.83} & 31.56 \\ \hline
			ALCNet                        & \multicolumn{1}{c|}{69.52} & \multicolumn{1}{c|}{\textbf{95.44}} & {47.13} & \multicolumn{1}{c|}{70.50} & \multicolumn{1}{c|}{95.66} & 13.79 & \multicolumn{1}{c|}{62.81} & \multicolumn{1}{c|}{89.56} & {29.26} 
			& \multicolumn{1}{c|}{67.61} & \multicolumn{1}{c|}{93.55} & 30.06 \\ \hline
			RDIAN                        & \multicolumn{1}{c|}{70.70} & \multicolumn{1}{c|}{94.30} &{29.22} & \multicolumn{1}{c|}{82.05} & \multicolumn{1}{c|}{97.25} & 12.94 & \multicolumn{1}{c|}{62.60} & \multicolumn{1}{c|}{86.87} & {19.59} 
			& \multicolumn{1}{c|}{71.78} & \multicolumn{1}{c|}{92.80} & 20.58 \\ \hline
			DNANet                        & \multicolumn{1}{c|}{76.61} & \multicolumn{1}{c|}{95.06} & {\underline{13.31}} & \multicolumn{1}{c|}{\underline{93.64}} & \multicolumn{1}{c|}{\textbf{98.94}} & {\underline{3.98}} & \multicolumn{1}{c|}{64.33} & \multicolumn{1}{c|}{89.56} &{\underline{11.67}}
			& \multicolumn{1}{c|}{78.19} & \multicolumn{1}{c|}{\underline{94.52}} & {\underline{9.65}}  \\ \hline
			
			ACM                        & \multicolumn{1}{c|}{67.09} & \multicolumn{1}{c|}{92.02} & 40.61 & \multicolumn{1}{c|}{65.90} & \multicolumn{1}{c|}{96.93} & 17.05 & \multicolumn{1}{c|}{62.45} & \multicolumn{1}{c|}{\underline{89.90}} & 46.67  & \multicolumn{1}{c|}{65.14} & \multicolumn{1}{c|}{92.95} & 34.78   \\ \hline
			ISNet                        & \multicolumn{1}{c|}{71.21} & \multicolumn{1}{c|}{93.16} & 46.31 & \multicolumn{1}{c|}{79.90} & \multicolumn{1}{c|}{97.57} & 14.20 & \multicolumn{1}{c|}{62.24} & \multicolumn{1}{c|}{89.23} & 25.51  & \multicolumn{1}{c|}{71.12} & \multicolumn{1}{c|}{93.32} & 28.67   \\ \hline
			Ours                      & \multicolumn{1}{c|}{\underline{77.38}} & \multicolumn{1}{c|}{\textbf{95.44}} & {19.96} & \multicolumn{1}{c|}{\textbf{94.53}} & \multicolumn{1}{c|}{\underline{98.52}} & {\textbf{1.52}} & \multicolumn{1}{c|}{\underline{64.70}} & \multicolumn{1}{c|}{\textbf{90.57}} &{39.17} & \multicolumn{1}{c|}{\underline{78.87}} & \multicolumn{1}{c|}{\textbf{94.84}} & 20.22 \\ \hline
			Ours$^{*}$                        & \multicolumn{1}{c|}{\textbf{77.88}} & \multicolumn{1}{c|}{\textbf{95.44}} & {31.35} & \multicolumn{1}{c|}{93.46} & \multicolumn{1}{c|}{98.41} & 4.46 & \multicolumn{1}{c|}{\textbf{67.53}} & \multicolumn{1}{c|}{89.56} & {21.05} & \multicolumn{1}{c|}{\textbf{79.62}} & \multicolumn{1}{c|}{94.47} & 18.95  \\ \hline
		\end{tabular}
	}
	\vspace{-0.2cm}
	\caption{Numerical results compared with a series of advanced methods on three representative datasets.}~\label{tab:numerical_general}
\end{table*}
\begin{table*}[htb]
	\centering
	\footnotesize
	\renewcommand{\arraystretch}{1.1}
	\setlength{\tabcolsep}{0.28mm}{
		\begin{tabular}{|c|cc|cc|cc|cc|cc|cc|cc|cc|cc|cc|}
			\hline
			\multirow{2}{*}{Methods} & \multicolumn{2}{c|}{\makecell[c]{\footnotesize  Gaussian  \\ \footnotesize  Noise}} & \multicolumn{2}{c|}{\makecell[c]{\footnotesize  Shot  \\ \footnotesize  Noise}} & \multicolumn{2}{c|}{\makecell[c]{\footnotesize  Defocus  \\\footnotesize  Blur}} & \multicolumn{2}{c|}{\makecell[c]{\footnotesize  Motion   \\ \footnotesize  Blur}} & \multicolumn{2}{c|}{\makecell[c]{\footnotesize  Gaussian   \\ \footnotesize  Blur}}& \multicolumn{2}{c|}{\footnotesize  Brightness}  & \multicolumn{2}{c|}{\footnotesize Contrast}  & \multicolumn{2}{c|}{\footnotesize  \footnotesize  Pixelate}  & \multicolumn{2}{c|}{\makecell[c]{\footnotesize  JPEG   \\ \footnotesize  Compression}} & \multicolumn{2}{c|}{\footnotesize Average}   \\ \cline{2-21} 
			& \multicolumn{1}{c|}{\cellcolor{gray!20}\footnotesize IOU$\uparrow$}    & {\cellcolor{gray!20}\footnotesize RCE$\downarrow$}   & \multicolumn{1}{c|}{\cellcolor{gray!20}\footnotesize IOU$\uparrow$}  & {\cellcolor{gray!20}\footnotesize RCE$\downarrow$} & \multicolumn{1}{c|}{\cellcolor{gray!20}\footnotesize IOU$\uparrow$}  & {\cellcolor{gray!20}\footnotesize RCE$\downarrow$}  & \multicolumn{1}{c|}{\cellcolor{gray!20}\footnotesize IOU$\uparrow$}   & {\cellcolor{gray!20}\footnotesize RCE$\downarrow$}   & \multicolumn{1}{c|}{\cellcolor{gray!20}\footnotesize IOU$\uparrow$}   & {\cellcolor{gray!20}\footnotesize RCE$\downarrow$}  & \multicolumn{1}{c|}{\cellcolor{gray!20}\footnotesize IOU$\uparrow$} & {\cellcolor{gray!20}\footnotesize RCE$\downarrow$} & \multicolumn{1}{c|}{\cellcolor{gray!20}\footnotesize IOU$\uparrow$} & {\cellcolor{gray!20}\footnotesize RCE$\downarrow$} & \multicolumn{1}{c|}{\cellcolor{gray!20}\footnotesize IOU$\uparrow$}     & {\cellcolor{gray!20}\footnotesize RCE$\downarrow$}    & \multicolumn{1}{c|}{\cellcolor{gray!20}\footnotesize IOU$\uparrow$}  & {\cellcolor{gray!20}\footnotesize RCE$\downarrow$} & \multicolumn{1}{c|}{\cellcolor{gray!20}\footnotesize IOU$\uparrow$} & {\cellcolor{gray!20}\footnotesize RCE$\downarrow$} \\ \hline
			ALCNet	& \multicolumn{1}{c|}{\underline{18.16}}       & 73.46       & \multicolumn{1}{c|}{\underline{18.89}}     & 72.38     & \multicolumn{1}{c|}{37.60}     & 45.03     & \multicolumn{1}{c|}{28.96}      & 57.67       & \multicolumn{1}{c|}{\underline{44.97}}      & 34.26      & \multicolumn{1}{c|}{54.49}    & 20.34    & \multicolumn{1}{c|}{38.93}    & 43.09     & \multicolumn{1}{c|}{{\textbf{63.61}}}     & 7.01    & \multicolumn{1}{c|}{59.46}        & 13.07       & \multicolumn{1}{c|}{40.56}    &  40.71   \\ \hline
			RDIAN	& \multicolumn{1}{c|}{13.33}       & 80.73      & \multicolumn{1}{c|}{11.21}     &   83.79  & \multicolumn{1}{c|}{30.98}     &   55.20   & \multicolumn{1}{c|}{24.02}      &  65.27     & \multicolumn{1}{c|}{38.42}      &  44.45    & \multicolumn{1}{c|}{54.94}    & 20.55    & \multicolumn{1}{c|}{32.52}    &  52.97   & \multicolumn{1}{c|}{59.74}     & 13.61    & \multicolumn{1}{c|}{58.40}        & 15.56     & \multicolumn{1}{c|}{35.95}    &   48.02  \\ \hline
			ACM 	& \multicolumn{1}{c|}{{15.56}}       &  {77.29}     & \multicolumn{1}{c|}{{15.65}}     &  {77.16}   & \multicolumn{1}{c|}{{35.75}}     & {47.82}      & \multicolumn{1}{c|}{{27.77}}      & {59.48}       & \multicolumn{1}{c|}{{43.08}}      &  {37.13}    & \multicolumn{1}{c|}{{55.58}}    &   {18.89}  & \multicolumn{1}{c|}{{35.09}}    &  {48.79}   & \multicolumn{1}{c|}{\underline{63.34}}        & {7.56}       & \multicolumn{1}{c|}{\underline{61.05}}     & {10.91}    & \multicolumn{1}{c|}{{39.21}}    & {42.78}    \\ \hline

			DNANet	& \multicolumn{1}{c|}{{17.40}}      & {77.64}      & \multicolumn{1}{c|}{{17.51}}     & {77.50}    & \multicolumn{1}{c|}{{32.50}}     & {58.24}     & \multicolumn{1}{c|}{\underline{29.56}}      &  {62.01}    & \multicolumn{1}{c|}{{43.15}}      &  {44.54}    & \multicolumn{1}{c|}{\underline{61.67}}    & {20.75}     & \multicolumn{1}{c|}{{45.76}}    &  {41.20}   & \multicolumn{1}{c|}{{62.38}}        & {19.84}        & \multicolumn{1}{c|}{{\textbf{63.10}}}     &  {18.91}   & \multicolumn{1}{c|}{{41.45}}    & {46.74}    \\ \hline

			ISNet 	& \multicolumn{1}{c|}{{16.07}}      & {77.19}      & \multicolumn{1}{c|}{{17.01}}     & {75.86}    & \multicolumn{1}{c|}{{\textbf{38.68}}}     & {45.12}     & \multicolumn{1}{c|}{{28.96}}      &  {58.90}    & \multicolumn{1}{c|}{{\textbf{47.52}}}      &  {32.57}    & \multicolumn{1}{c|}{{56.53}}    & {19.78}     & \multicolumn{1}{c|}{{\textbf{49.88}}}    &  {29.22}   & \multicolumn{1}{c|}{{61.27}  }        & {13.07}        & \multicolumn{1}{c|}{{59.12}}     &  {16.12}   & \multicolumn{1}{c|}{{\underline{41.67}}}    &   {40.88}  \\ \hline
			
			Ours	    & \multicolumn{1}{c|}{{\textbf{26.03}}}     &   66.58  & \multicolumn{1}{c|}{{\textbf{22.60}}}     &   70.99   & \multicolumn{1}{c|}{\underline{37.73}}      &  51.57     & \multicolumn{1}{c|}{{\textbf{30.89}}}      &  60.34    & \multicolumn{1}{c|}{42.90}    & 44.92    & \multicolumn{1}{c|}{{\textbf{61.81}}}    &  20.65   & \multicolumn{1}{c|}{\underline{46.50}}        &   40.31     & \multicolumn{1}{c|}{61.89}     & 20.55   & \multicolumn{1}{c|}{59.20}    & 24.00 & \multicolumn{1}{c|}{{\textbf{43.28}}}       & 44.44    \\ \hline
		\end{tabular}
	}
	\vspace{-0.2cm}
	\caption{Numerical results about the robustness of advanced learning-based methods on diverse corruption factors.}~\label{tab:corrup}
\end{table*}
\begin{table*}[htb]
	\centering
	\footnotesize
	\renewcommand{\arraystretch}{1.1}
	\setlength{\tabcolsep}{0.28mm}{
		\begin{tabular}{|c|cc|cc|cc|cc|cc|cc|cc|cc|cc|cc|}
			\hline
			\multirow{2}{*}{Methods} & \multicolumn{2}{c|}{\makecell[c]{\footnotesize  Gaussian  \\ \footnotesize  Noise}} & \multicolumn{2}{c|}{\makecell[c]{\footnotesize  Shot  \\ \footnotesize  Noise}} & \multicolumn{2}{c|}{\makecell[c]{\footnotesize  Defocus  \\\footnotesize  Blur}} & \multicolumn{2}{c|}{\makecell[c]{\footnotesize  Motion   \\ \footnotesize  Blur}} & \multicolumn{2}{c|}{\makecell[c]{\footnotesize  Gaussian   \\ \footnotesize  Blur}}& \multicolumn{2}{c|}{\footnotesize  Brightness}  & \multicolumn{2}{c|}{\footnotesize Contrast}  & \multicolumn{2}{c|}{\footnotesize  \footnotesize  Pixelate}  & \multicolumn{2}{c|}{\makecell[c]{\footnotesize  JPEG   \\ \footnotesize  Compression}} & \multicolumn{2}{c|}{\footnotesize Average}   \\ \cline{2-21} 
			& \multicolumn{1}{c|}{\cellcolor{gray!20}\footnotesize IOU$\uparrow$}    & {\cellcolor{gray!20}\footnotesize RCE$\downarrow$}   & \multicolumn{1}{c|}{\cellcolor{gray!20}\footnotesize IOU$\uparrow$}  & {\cellcolor{gray!20}\footnotesize RCE$\downarrow$} & \multicolumn{1}{c|}{\cellcolor{gray!20}\footnotesize IOU$\uparrow$}  & {\cellcolor{gray!20}\footnotesize RCE$\downarrow$}  & \multicolumn{1}{c|}{\cellcolor{gray!20}\footnotesize IOU$\uparrow$}   & {\cellcolor{gray!20}\footnotesize RCE$\downarrow$}   & \multicolumn{1}{c|}{\cellcolor{gray!20}\footnotesize IOU$\uparrow$}   & {\cellcolor{gray!20}\footnotesize RCE$\downarrow$}  & \multicolumn{1}{c|}{\cellcolor{gray!20}\footnotesize IOU$\uparrow$} & {\cellcolor{gray!20}\footnotesize RCE$\downarrow$} & \multicolumn{1}{c|}{\cellcolor{gray!20}\footnotesize IOU$\uparrow$} & {\cellcolor{gray!20}\footnotesize RCE$\downarrow$} & \multicolumn{1}{c|}{\cellcolor{gray!20}\footnotesize IOU$\uparrow$}     & {\cellcolor{gray!20}\footnotesize RCE$\downarrow$}    & \multicolumn{1}{c|}{\cellcolor{gray!20}\footnotesize IOU$\uparrow$}  & {\cellcolor{gray!20}\footnotesize RCE$\downarrow$} & \multicolumn{1}{c|}{\cellcolor{gray!20}\footnotesize IOU$\uparrow$} & {\cellcolor{gray!20}\footnotesize RCE$\downarrow$} \\ \hline
			ACM	& \multicolumn{1}{c|}{0.05}       & 99.93       & \multicolumn{1}{c|}{1.22}     & 98.19     & \multicolumn{1}{c|}{23.87}     & 64.43     & \multicolumn{1}{c|}{27.45}      & 59.09       & \multicolumn{1}{c|}{31.40}      & 53.21      & \multicolumn{1}{c|}{50.91}    & 24.13    & \multicolumn{1}{c|}{29.24}    &56.43     & \multicolumn{1}{c|}{59.79}        & 10.90       & \multicolumn{1}{c|}{60.13}     & 10.39    & \multicolumn{1}{c|}{31.56}    &  52.96   \\ \hline
			
			ACM$_\mathtt{P}$ 	& \multicolumn{1}{c|}{\textbf{15.56}}       &  \textbf{77.29}     & \multicolumn{1}{c|}{\textbf{15.65}}     &  \textbf{77.16}   & \multicolumn{1}{c|}{\textbf{35.75}}     & \textbf{47.82}      & \multicolumn{1}{c|}{\textbf{27.77}}      & \textbf{59.48}       & \multicolumn{1}{c|}{\textbf{43.08}}      &  \textbf{37.13}    & \multicolumn{1}{c|}{\textbf{55.58}}    &   \textbf{18.89}  & \multicolumn{1}{c|}{\textbf{35.09}}    &  \textbf{48.79}   & \multicolumn{1}{c|}{\textbf{63.34}}        & \textbf{7.56}       & \multicolumn{1}{c|}{\textbf{61.05}}     & \textbf{10.91}    & \multicolumn{1}{c|}{\textbf{39.21}}    & \textbf{42.78}    \\ \hline
			DNA	& \multicolumn{1}{c|}{2.81}       & 96.33      & \multicolumn{1}{c|}{2.66}     &   96.53  & \multicolumn{1}{c|}{24.93}     &   67.46   & \multicolumn{1}{c|}{27.00}      &  64.76     & \multicolumn{1}{c|}{33.78}      &  55.91    & \multicolumn{1}{c|}{58.91}    & 23.12    & \multicolumn{1}{c|}{30.40}    &  60.32   & \multicolumn{1}{c|}{57.58}        &   24.86     & \multicolumn{1}{c|}{61.88}     & 19.24    & \multicolumn{1}{c|}{33.33}    &   56.50  \\ \hline
			
			DNA$_\mathtt{P}$ 	& \multicolumn{1}{c|}{\textbf{17.40}}      & \textbf{77.64}      & \multicolumn{1}{c|}{\textbf{17.51}}     & \textbf{77.50}    & \multicolumn{1}{c|}{\textbf{32.50}}     & \textbf{58.24}     & \multicolumn{1}{c|}{\textbf{29.56}}      &  \textbf{62.01}    & \multicolumn{1}{c|}{\textbf{43.15}}      &  \textbf{44.54}    & \multicolumn{1}{c|}{\textbf{61.67}}    & \textbf{20.75}     & \multicolumn{1}{c|}{\textbf{45.76}}    &  \textbf{41.20}   & \multicolumn{1}{c|}{\textbf{62.38}  }        & \textbf{19.84}        & \multicolumn{1}{c|}{\textbf{63.10}}     &  \textbf{18.91}   & \multicolumn{1}{c|}{\textbf{41.45}}    & \textbf{46.74}    \\ \hline
			
			ISNet	    & \multicolumn{1}{c|}{0.30}     &   99.58  & \multicolumn{1}{c|}{0.98}     &   98.62   & \multicolumn{1}{c|}{22.01}      &  69.10     & \multicolumn{1}{c|}{27.78}      &  60.99    & \multicolumn{1}{c|}{31.75}    & 55.41    & \multicolumn{1}{c|}{54.80}    &  23.04   & \multicolumn{1}{c|}{32.83}        &   53.90     & \multicolumn{1}{c|}{58.81}     & 17.42   & \multicolumn{1}{c|}{59.18}    & 16.90 & \multicolumn{1}{c|}{29.84}       & 58.09    \\ \hline
			
			ISNet$_\mathtt{P}$ 	& \multicolumn{1}{c|}{\textbf{16.07}}      & \textbf{77.19}      & \multicolumn{1}{c|}{\textbf{17.01}}     & \textbf{75.86}    & \multicolumn{1}{c|}{\textbf{38.68}}     & \textbf{45.12}     & \multicolumn{1}{c|}{\textbf{28.96}}      &  \textbf{58.90}    & \multicolumn{1}{c|}{\textbf{47.52}}      &  \textbf{32.57}    & \multicolumn{1}{c|}{\textbf{56.53}}    & \textbf{19.78}     & \multicolumn{1}{c|}{\textbf{49.88}}    &  \textbf{29.22}   & \multicolumn{1}{c|}{\textbf{61.27}  }        & \textbf{13.07}        & \multicolumn{1}{c|}{\textbf{59.12}}     &  \textbf{16.12}   & \multicolumn{1}{c|}{\textbf{41.67}}    &   \textbf{40.88}  \\ \hline
		\end{tabular}
	}
	\vspace{-0.2cm}
	\caption{Evaluating the effectiveness of the proposed training strategy with diverse corruptions on the NUAA dataset.}~\label{tab:generalization2}
\end{table*}

\begin{table*}[htb]
	\centering
	\footnotesize
	\renewcommand{\arraystretch}{1.1}
	\setlength{\tabcolsep}{0.28mm}{
		\begin{tabular}{|c|cc|cc|cc|cc|cc|cc|cc|cc|cc|cc|}
			\hline
			\multirow{2}{*}{Methods} & \multicolumn{2}{c|}{\makecell[c]{\footnotesize  Gaussian  \\ \footnotesize  Noise}} & \multicolumn{2}{c|}{\makecell[c]{\footnotesize  Shot  \\ \footnotesize  Noise}} & \multicolumn{2}{c|}{\makecell[c]{\footnotesize  Defocus  \\\footnotesize  Blur}} & \multicolumn{2}{c|}{\makecell[c]{\footnotesize  Motion   \\ \footnotesize  Blur}} & \multicolumn{2}{c|}{\makecell[c]{\footnotesize  Gaussian   \\ \footnotesize  Blur}}& \multicolumn{2}{c|}{\footnotesize  Brightness}  & \multicolumn{2}{c|}{\footnotesize Contrast}  & \multicolumn{2}{c|}{\footnotesize  \footnotesize  Pixelate}  & \multicolumn{2}{c|}{\makecell[c]{\footnotesize  JPEG   \\ \footnotesize  Compression}} & \multicolumn{2}{c|}{\footnotesize Average}   \\ \cline{2-21} 
			& \multicolumn{1}{c|}{\cellcolor{gray!20}\footnotesize IOU$\uparrow$}    & {\cellcolor{gray!20}\footnotesize RCE$\downarrow$}   & \multicolumn{1}{c|}{\cellcolor{gray!20}\footnotesize IOU$\uparrow$}  & {\cellcolor{gray!20}\footnotesize RCE$\downarrow$} & \multicolumn{1}{c|}{\cellcolor{gray!20}\footnotesize IOU$\uparrow$}  & {\cellcolor{gray!20}\footnotesize RCE$\downarrow$}  & \multicolumn{1}{c|}{\cellcolor{gray!20}\footnotesize IOU$\uparrow$}   & {\cellcolor{gray!20}\footnotesize RCE$\downarrow$}   & \multicolumn{1}{c|}{\cellcolor{gray!20}\footnotesize IOU$\uparrow$}   & {\cellcolor{gray!20}\footnotesize RCE$\downarrow$}  & \multicolumn{1}{c|}{\cellcolor{gray!20}\footnotesize IOU$\uparrow$} & {\cellcolor{gray!20}\footnotesize RCE$\downarrow$} & \multicolumn{1}{c|}{\cellcolor{gray!20}\footnotesize IOU$\uparrow$} & {\cellcolor{gray!20}\footnotesize RCE$\downarrow$} & \multicolumn{1}{c|}{\cellcolor{gray!20}\footnotesize IOU$\uparrow$}     & {\cellcolor{gray!20}\footnotesize RCE$\downarrow$}    & \multicolumn{1}{c|}{\cellcolor{gray!20}\footnotesize IOU$\uparrow$}  & {\cellcolor{gray!20}\footnotesize RCE$\downarrow$} & \multicolumn{1}{c|}{\cellcolor{gray!20}\footnotesize IOU$\uparrow$} & {\cellcolor{gray!20}\footnotesize RCE$\downarrow$} \\ \hline
			ACM	& \multicolumn{1}{c|}{0.26}       & 99.61       & \multicolumn{1}{c|}{3.03}     & 98.74     & \multicolumn{1}{c|}{8.78}     & 86.67     & \multicolumn{1}{c|}{26.59}      & 59.65       & \multicolumn{1}{c|}{16.09}      & 75.58      & \multicolumn{1}{c|}{50.96}    & 22.66    & \multicolumn{1}{c|}{32.52}    &50.65    & \multicolumn{1}{c|}{42.25}        & 35.89       & \multicolumn{1}{c|}{39.95}     & 39.37    & \multicolumn{1}{c|}{24.83}    &63.20     \\ \hline
			
			ACM$_\mathtt{P}$ 	& \multicolumn{1}{c|}{\textbf{3.88}}       &  \textbf{94.15}     & \multicolumn{1}{c|}{\textbf{5.25}}     &  \textbf{92.09}   & \multicolumn{1}{c|}{\textbf{24.09}}     & \textbf{63.70}      & \multicolumn{1}{c|}{\textbf{30.40}}      & \textbf{54.19}       & \multicolumn{1}{c|}{\textbf{29.09}}      &  \textbf{56.15}    & \multicolumn{1}{c|}{\textbf{52.52}}    &   \textbf{20.86}  & \multicolumn{1}{c|}{\textbf{48.04}}    &  \textbf{27.61}   & \multicolumn{1}{c|}{\textbf{48.69}}        & \textbf{26.62}       & \multicolumn{1}{c|}{\textbf{41.75}}     & \textbf{37.09}    & \multicolumn{1}{c|}{\textbf{31.52}}    & \textbf{52.50}    \\ \hline
			DNA	& \multicolumn{1}{c|}{0.12}       & 99.88      & \multicolumn{1}{c|}{0.60}     &   99.36  & \multicolumn{1}{c|}{7.96}     &   91.49   & \multicolumn{1}{c|}{30.77}      &  67.14     & \multicolumn{1}{c|}{18.62}      &  80.11    & \multicolumn{1}{c|}{60.43}    & 35.46    & \multicolumn{1}{c|}{54.44}    &  41.56   & \multicolumn{1}{c|}{33.40}        &   64.33     & \multicolumn{1}{c|}{24.90}     & 73.40    & \multicolumn{1}{c|}{25.69}    &  72.53   \\ \hline
			
			DNA$_\mathtt{P}$	& \multicolumn{1}{c|}{\textbf{9.12}}      & \textbf{90.26}      & \multicolumn{1}{c|}{\textbf{8.08}}     & \textbf{91.38}    & \multicolumn{1}{c|}{\textbf{30.45}}     & \textbf{67.50}     & \multicolumn{1}{c|}{\textbf{40.24}}      &  \textbf{57.05}    & \multicolumn{1}{c|}{\textbf{33.78}}      &  \textbf{63.95}    & \multicolumn{1}{c|}{\textbf{72.10}}    & \textbf{23.05}     & \multicolumn{1}{c|}{\textbf{75.26}}    &  \textbf{19.68}   & \multicolumn{1}{c|}{\textbf{57.08}  }        & \textbf{39.09}        & \multicolumn{1}{c|}{\textbf{38.16}}     &  \textbf{59.28}   & \multicolumn{1}{c|}{\textbf{40.47}}    &    \textbf{56.80} \\ \hline
			
			ISNet	    & \multicolumn{1}{c|}{0.17}     &   99.78  & \multicolumn{1}{c|}{1.16}     &   98.55   & \multicolumn{1}{c|}{3.31}      &  95.86     & \multicolumn{1}{c|}{27.58}      &  65.48    & \multicolumn{1}{c|}{31.75}    & 55.41    & \multicolumn{1}{c|}{54.80}    &  23.04   & \multicolumn{1}{c|}{32.83}        &   53.90     & \multicolumn{1}{c|}{58.81}     & 17.42   & \multicolumn{1}{c|}{59.18}    & 16.90 & \multicolumn{1}{c|}{29.95}       & 58.48   \\ \hline
			
			ISNet$_\mathtt{P}$ 	& \multicolumn{1}{c|}{\textbf{4.77}}      & \textbf{94.03}      & \multicolumn{1}{c|}{\textbf{6.15}}     & \textbf{92.09}    & \multicolumn{1}{c|}{\textbf{30.42}}     & \textbf{61.92}     & \multicolumn{1}{c|}{\textbf{34.03}}      &  \textbf{57.41}    & \multicolumn{1}{c|}{\textbf{29.53}}      &  \textbf{63.04}    & \multicolumn{1}{c|}{\textbf{59.42}}    & \textbf{25.63}     & \multicolumn{1}{c|}{\textbf{56.90}}    &  \textbf{28.78}   & \multicolumn{1}{c|}{\textbf{50.89}  }        & \textbf{36.31}        & \multicolumn{1}{c|}{\textbf{42.00}}     &  \textbf{47.43}   & \multicolumn{1}{c|}{\textbf{34.90}}    &   \textbf{56.29}  \\ \hline
			
		\end{tabular}
	}
	\vspace{-0.2cm}
	\caption{Evaluating the effectiveness of proposed training strategy with diverse corruptions on the NUDT dataset.}~\label{tab:generalization3}
\end{table*}
\section{Experiments}

\begin{figure*}[!htb]
	\centering
	\setlength{\tabcolsep}{1pt} 
	
	\includegraphics[width=0.99\textwidth,height=0.24\textheight]{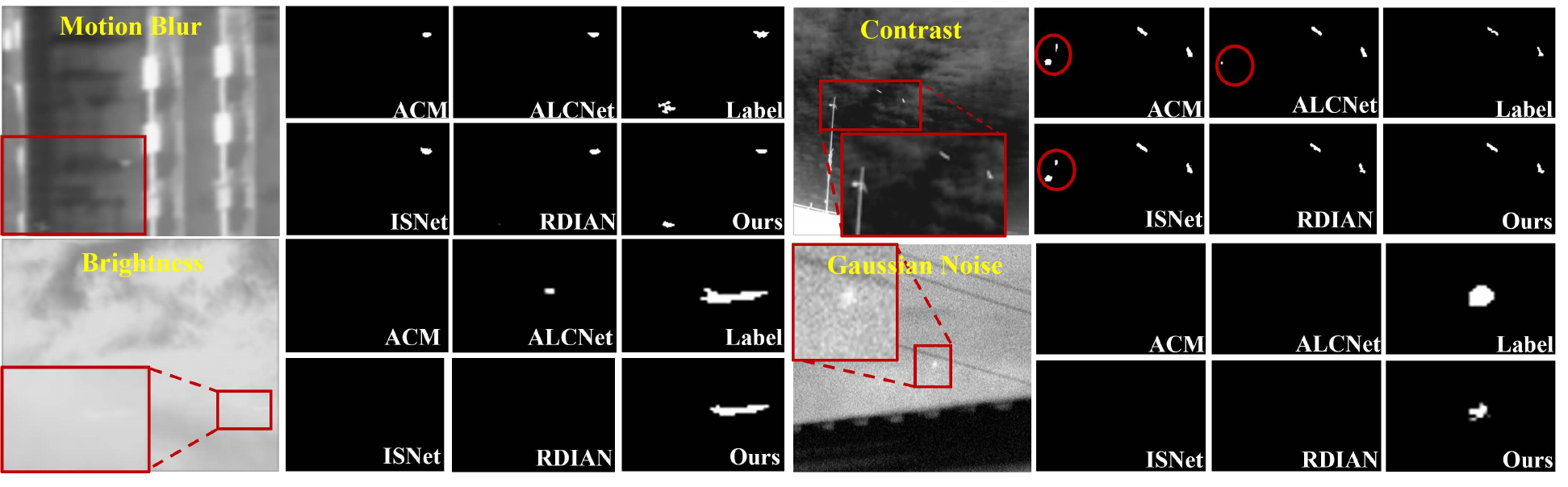}
	
	\caption{Qualitative comparisons with four advanced competitors under four kinds of corruptions.}
	\label{fig:corruption}
\end{figure*}
\subsection{Implementation Details}

\noindent \textbf{Corruptions in infrared imaging.} To simulate real-world corruptions, we leverage 10 different perturbation strategies with 3 levels of severity, which can be divided into four categories: (1) Noise: which includes Gaussian noise, shot noise, and impulse noise; (2) Blur: which includes motion blur and defocus blur; (3) ISP simulating: which includes brightness, contrast, pixelate and JPEG compression. Noting that we do not consider the weather, where infrared imaging is almost not sensitive to illumination changes and weathers. 

\noindent \textbf{Datasets and evaluation metrics.} As for the datasets, we utilize three representative benchmarks to train and evaluate our algorithms, including   NUAA~\cite{dai2021attentional}, NUDT~\cite{li2022dense}, and IRSTD-1K~\cite{zhang2022isnet}. Following the practice~\cite{ying2023mapping}, we split the training and testing sets in the same way.

As for the metrics, we utilize three kinds of criteria, including Intersection Over Union (IOU) to measure pixel-wise accuracy,
Probability of detection (P$_d$) and False-alarm rate (F$_a$) to gauge target-wise precision. In order to measure the robustness of corruption, we also introduce the Relative  Corruption Error (RCE)~\cite{dong2023benchmarking}, which is defined as:
\begin{equation}\label{eq:rce}
RCE = \frac{\text{IOU}_\mathtt{clean}-\text{IOU}_\mathtt{cor}}{\text{IOU}_\mathtt{clean}},
\end{equation}
where $\text{IOU}_\mathtt{clean}$ and $\text{IOU}_\mathtt{cor}$ are the measurements of clean and corrupted datasets respectively.

\noindent \textbf{Training configurations.}  We leveraged the Adam~\cite{kingma2014adam} and SGD optimizers to train $\mathcal{N}_\mathtt{D}$ and $\mathcal{N}_{S}$ with initial learning rates $5e^{-4} $ and $1e^{-4} $ respectively. Soft-IOU loss is the criterion (\textit{i.e.,} $\mathcal{L}$) and $\lambda = 1$.
Data augmentation, such as randomly flipping and cropping are implemented for training with patches of size $256 \times 256$.
All experiments were implemented in PyTorch with an Nvidia Tesla V100 GPU. 
We compared with ten state-of-arts methods, including  traditional approaches, \textit{i.e.,} Top-Hat~\cite{rivest1996detection}, MSPCM~\cite{moradi2018false}, NRAM~\cite{zhang2018infrared}, IPI~\cite{gao2013infrared} and PSTNN~\cite{zhang2019infrared} and learning-based schemes including ALCNet~\cite{dai2021attentional}, RDIAN~\cite{sun2023receptive}, DNANet~\cite{li2022dense}, ACM~\cite{dai2021asymmetric}, ISNet~\cite{zhang2022isnet}.

\vspace{-1em}
\subsection{Results on Standard Benchmarks}

\noindent \textbf{Quantitative results.} We report the numerical comparisons with ten advanced compositors on three representative benchmarks, which are shown in Table.~\ref{tab:numerical_general}.  We provide two variants of our scheme, where ``Ours'' denotes the general version of joint training and ``Ours$^{*}$'' represents the model based on the adversarial training. Our schemes realize the consistently remarkable performance on all three benchmarks in terms of IOU, which reflects our methods can better detect the informative characteristics, such as target shape and texture edges. Note that IRSTD-1K is a more challenging dataset compared to the previous ones, containing obvious clutters and noises in the background with degraded targets (various shapes with low contrast). Our method can drastically improve 4.97\% of IOU than advanced DNANet and realize promising results  (1.83\% averaged promotion).

\noindent \textbf{Qualitative results.} Figure.~\ref{fig:compare} depicts the visual comparisons with typical learning-based schemes on three challenging scenarios (\textit{i.e.,} varied shapes under complicated background, low contrast, and low SCR). Obviously, our scheme has three significant advantages. First, our approach can effectively extract the edges of targets from the complex background. For instance, as shown in the first row, the small infrared target is camouflaged in dense bushes with abundant texture details. 
Our method can estimate the accurate shape from the messy background. Secondly, our method can avoid the interferences of confused objects, shown in the second row of  Figure.~\ref{fig:compare}. 
Most methods incorrectly predict some clouds as small infrared targets. Benefiting from the adversarial training under diver contrast degrees and 
effective frequency refinement, our approaches still preserve the curial shape of targets, removing the distractions of confused objects. Lastly, our method can discover the target with precise shape estimation under extremely challenging scenarios. The complete shape of the low-SCR infrared target can be accurately estimated, shown in the third row. 

\begin{table*}[htb]
	\centering
	\footnotesize
	\renewcommand{\arraystretch}{1.1}
	\setlength{\tabcolsep}{1.1mm}{
		\begin{tabular}{|c|ccc|ccc|ccc|}
			\hline
			\multirow{2}{*}{Methods} & \multicolumn{3}{c|}{NUAA}                        & \multicolumn{3}{c|}{NUDT}                            & \multicolumn{3}{c|}{IRSTD-1K}                            \\ \cline{2-10} 
			
			& \multicolumn{1}{c|}{\cellcolor{gray!20}IOU$\uparrow$} & \multicolumn{1}{c|}{\cellcolor{gray!20}P$_{d}$$\uparrow$} & \multicolumn{1}{c|}{\cellcolor{gray!20}F$_{a}$$\downarrow$} & \multicolumn{1}{c|}{\cellcolor{gray!20}IOU$\uparrow$} & \multicolumn{1}{c|}{\cellcolor{gray!20}P$_{d}$$\uparrow$} & \multicolumn{1}{c|}{\cellcolor{gray!20}F$_{a}$$\downarrow$}& \multicolumn{1}{c|}{\cellcolor{gray!20}IOU$\uparrow$} & \multicolumn{1}{c|}{\cellcolor{gray!20}P$_{d}$$\uparrow$} & \multicolumn{1}{c|}{\cellcolor{gray!20}F$_{a}$$\downarrow$}  \\ \hline
			ACM                        & \multicolumn{1}{c|}{67.09} & \multicolumn{1}{c|}{92.02} & 40.61 & \multicolumn{1}{c|}{65.90} & \multicolumn{1}{c|}{\textbf{96.93}} & 17.05 & \multicolumn{1}{c|}{62.45} & \multicolumn{1}{c|}{89.90} & 46.71  \\ \hline
			ACM$_\mathtt{P}$                        & \multicolumn{1}{c|}{\textbf{68.53}$_{\uparrow 2.15\%}$} & \multicolumn{1}{c|}{\textbf{92.40}$_{\uparrow 0.41\%}$} & \textbf{38.27}$_{\downarrow 5.76\%}$ & \multicolumn{1}{c|}{\textbf{66.37}$_{\uparrow 0.71\%}$} & \multicolumn{1}{c|}{95.56} & \textbf{14.94}$_{\downarrow 12.38\%}$ & \multicolumn{1}{c|}{\textbf{63.25}$_{\uparrow 1.28\%}$} & \multicolumn{1}{c|}{\textbf{90.24}$_{\uparrow 0.38\%}$} & \textbf{34.14}$_{\downarrow 26.91\%}$  \\ \hline
			DNA                        & \multicolumn{1}{c|}{76.61} & \multicolumn{1}{c|}{95.06} & \textbf{13.31} & \multicolumn{1}{c|}{93.64} & \multicolumn{1}{c|}{98.94} & 3.98 & \multicolumn{1}{c|}{64.33} & \multicolumn{1}{c|}{\textbf{89.56}} & 11.67  \\ \hline
			DNA$_\mathtt{P}$                        & \multicolumn{1}{c|}{\textbf{77.83}$_{\uparrow 1.59\%}$} & \multicolumn{1}{c|}{\textbf{96.20}$_{\uparrow 1.20\%}$} & 15.37 & \multicolumn{1}{c|}{\textbf{93.70}$_{\uparrow 0.064\%}$} & \multicolumn{1}{c|}{\textbf{99.26}$_{\uparrow 0.32\%}$} & \textbf{3.17}$_{\downarrow 20.4\%}$ & \multicolumn{1}{c|}{\textbf{65.28}$_{\uparrow 1.48\%}$} & \multicolumn{1}{c|}{89.23} & \textbf{7.12}$_{\downarrow 39.0\% }$ \\ \hline
		\end{tabular}
	}
	\vspace{-0.2cm}
	\caption{Evaluating the generalization ability of proposed training strategy on three general datasets.}~\label{tab:generalization}
\end{table*}

\subsection{Robustness on Corrupted Scenarios}

\noindent \textbf{Quantitative results.} Table.~\ref{tab:corrup} reports the robustness comparisons with five advanced learning-based approaches to defend corruptions on the NUAA dataset.  All these learning-based schemes are retrained under our bi-level adversarial framework. Due to the significant feature refinement ability of SFIM, our scheme realize the 3.86\%  averaged promotions compared with these advanced competitors, which demonstrates the effectiveness of our scheme to improve robustness. Our methods are the robustest for diverse noise factors and motion blur, which is crucial for real-world applications.

\noindent \textbf{Qualitative results.} Figure.~\ref{fig:corruption} illustrates the visual comparison of diverse corruptions (\textit{i.e.,} motion blur, contrast, brightness, and Gaussian noise). Though these methods were also trained on our hierarchical reinforced learning strategy, they still have limitations of architecture, leading to missed and false detections. 
Our proposed scheme achieves remarkable performance under various corrupted scenes.
As shown in the case of motion blur, the blurred target under the low-contrast background can be precisely detected.  We also provide two severe conditions, which contain strong brightness and heavy noise. Most of the schemes failed to detect the small infrared targets. Because the proposed frequency refinement has the powerful ability to disentangle the salient features, our method achieves consistent performance.

\section{Ablation Studies}
\begin{table}[htb]
	\centering
	\footnotesize
	\renewcommand{\arraystretch}{1.1}
	\setlength{\tabcolsep}{2.5mm}{
		\begin{tabular}{ccccc}
			\cline{1-4}
			\multicolumn{1}{|c|}{Strategy}                  & \multicolumn{1}{c|}{IOU$_\mathtt{clean}\uparrow$} & \multicolumn{1}{c|}{IOU$_\mathtt{cor}\uparrow$} & \multicolumn{1}{c|}{RCE$\downarrow$}             \\ \cline{1-4}
			\multicolumn{1}{|c|}{Baseline}        & \multicolumn{1}{c|}{67.10}              & \multicolumn{1}{c|}{29.87}            & \multicolumn{1}{c|}{55.48}         \\ \cline{1-4}
			\multicolumn{1}{|c|}{Random}       & \multicolumn{1}{c|}{68.34$_{\uparrow 1.85\%}$}              & \multicolumn{1}{c|}{31.09$_{\uparrow 4.08\%}$}            & \multicolumn{1}{c|}{{54.51}} \\ \cline{1-4}
			\multicolumn{1}{|c|}{Nosie}        & \multicolumn{1}{c|}{67.81$_{\uparrow 1.06\%}$}              & \multicolumn{1}{c|}{36.17$_{\uparrow 21.09\%}$}            & \multicolumn{1}{c|}{\textbf{46.66}}         \\ \cline{1-4}
			\multicolumn{1}{|c|}{Blur}         & \multicolumn{1}{c|}{67.60$_{\uparrow 0.75\%}$}              & \multicolumn{1}{c|}{31.65$_{\uparrow 5.96\%}$}            & \multicolumn{1}{c|}{53.19}           \\ \cline{1-4}
			\multicolumn{1}{|c|}{ISP Degradation} & \multicolumn{1}{c|}{67.36$_{\uparrow 0.39\%}$}              & \multicolumn{1}{c|}{30.58$_{\uparrow 2.38\%}$}            & \multicolumn{1}{l|}{54.73}         \\ \cline{1-4}
			\multicolumn{1}{|c|}{Ours (HRL)}          & \multicolumn{1}{c|}{\textbf{68.52$_{\uparrow 2.12\%}$}}     & \multicolumn{1}{c|}{\textbf{36.43$_{\uparrow 21.96\%}$}}   & \multicolumn{1}{c|}{46.84}          \\ \cline{1-4}
			
		\end{tabular}
	}
	\vspace{-0.2cm}
	\caption{ Comparison with  different training strategies (random selection and one corruptions) on the NUAA dataset.}~\label{tab:ts}
\end{table}

\begin{figure}[htb]
	\centering
	\includegraphics[width=0.48\textwidth]{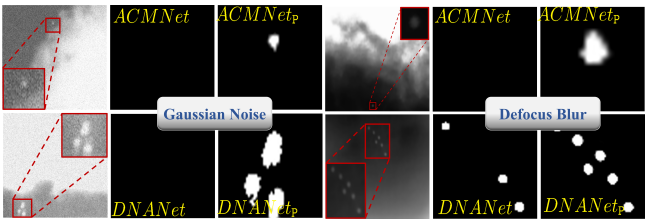}
	
	\caption{{Verification of the effectiveness of training strategies for different learnable networks under two kinds of corruptions \textit{i.e.,} Gaussian noise and defocus blur}.}
	\label{fig:generalization}
\end{figure}

\noindent \textbf{Effectiveness of training strategy.}  We compare the proposed training strategy HRL with random selection, and single corruption (\textit{i.e.,} noise, blur, and ISP degradation) in Table.~\ref{tab:ts}. We set the ACM as 
the baseline model. The random selection of strategy with cooperation training realizes better performance than ones under single corruption. Moreover, the learning of noise is significant for robustness.
Our strategy achieves a remarkable promotion, 2.12\% improvement on the general benchmark, and 21.96\% promotion on the corrupted scenes. 
We argue that our training strategy is network-agnostic, which can improve the arbitrary models of ISTD both for the general accuracy and robustness with corruptions. Table.~\ref{tab:generalization2} and Table.~\ref{tab:generalization3} report the numerical details of performance improvement leveraging the proposed HRL training strategy, where the subscript ``P'' denotes the proposed strategy. Under these nine degradation conditions, the performance of the three models is significantly improved. Especially, our training strategy can endow the strong robustness of ISTD model under noise interference. When the model trained on normal data encounters noise, the detection basically fails. The visual comparisons shown in Figure.~\ref{fig:generalization} also demonstrate this statement. Besides that, our model also can drastically improve the performance of the general datasets, which is reported in Table.~\ref{tab:generalization}. For instance, the ACM under our training strategy can significantly increase 2.15\% performance under the NUAA dataset.

\noindent \textbf{Benchmarking of natural corruptions.} Following the recent works~\cite{dong2023benchmarking,ren2022benchmarking}, providing the benchmarks of robustness under corruptions, we can find some valuable insights that may boost the development of ISTD.
(1) ISTD models are robust for the systematic degradations of infrared ISP (such as JPEG Compression and pixelate). (2) The noise corruptions (\textit{e.g.,} Gaussian and shot noise) are the most harmful to the ISTD model, which realizes almost 99\% RCE. (3) Defocus blur has a higher impact than motion blur.

\begin{figure}[htb]
	\centering
	\includegraphics[width=0.48\textwidth]{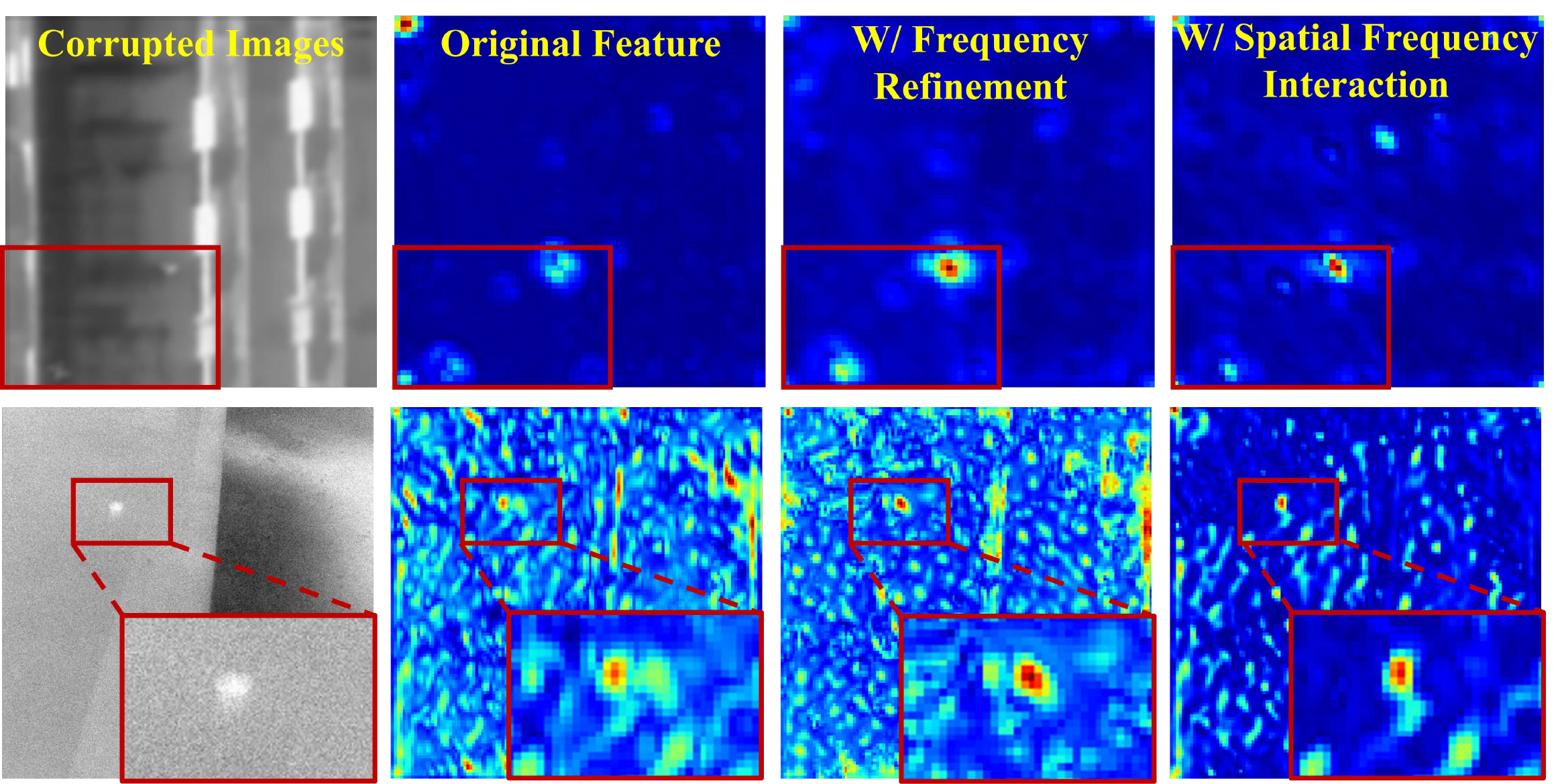}
	\vspace{-0.3cm}
	\caption{{ Feature visualization of different parts. From left to right: Degraded infrared images, original  features, features under frequency refinement and features after  SFIM}.}
	\label{fig:decom}
\end{figure}

\noindent \textbf{Impacts of spatial-frequency interaction.}  The proposed SFIM plays a key role in highlighting the salient features of corrupted scenes. Table.~\ref{tab:fi2} reports the quantitative results to demonstrate the effectiveness compared with the variant ``Ours$_\mathtt{w/o SFIM}$''. We also visualize the features of the procedure of SFIM under diverse corruptions (motion blur and Gaussian noise) in Figure.~\ref{fig:decom} The frequency refinement highlights the locations of thermal small targets. The spatial interaction can remove the degraded artefacts, shown in the second row of Figure.~\ref{fig:decom}.

\begin{table}[htb]
	
	\centering
	\footnotesize
	\renewcommand{\arraystretch}{1.1}
	\setlength{\tabcolsep}{2.5mm}{
		\begin{tabular}{cccc}
			\cline{1-4}
			\multicolumn{1}{|c|}{Model}        & \multicolumn{1}{c|}{IOU$\uparrow$}            & \multicolumn{1}{c|}{P$_{d}$$\uparrow$}             & \multicolumn{1}{c|}{F$_{a}$$\downarrow$}              \\ \cline{1-4}
			\multicolumn{1}{|c|}{Ours$_\mathtt{w/o SFIM}$} & \multicolumn{1}{c|}{75.02}          & \multicolumn{1}{c|}{93.92}          & \multicolumn{1}{c|}{37.66}           \\ \cline{1-4}
			\multicolumn{1}{|c|}{Ours}         & \multicolumn{1}{c|}{\textbf{77.38}$_{\uparrow 3.15\%}$} & \multicolumn{1}{c|}{\textbf{95.44}$_{\uparrow 1.62\%}$} & \multicolumn{1}{c|}{\textbf{19.96}$_{\downarrow 47.00\%}$}  \\ \cline{1-4}
		\end{tabular}
	}
	\vspace{-0.2cm}
	\caption{Effectiveness of spatial-frequency interaction module on the NUAA dataset.}~\label{tab:fi2}
\end{table}

\section{Conclusion}
In this paper, a bi-level adversarial framework was proposed to address the robustness of infrared small target detection models.  A hierarchical reinforced learning strategy was introduced to construct the competitive game to automatically discover the harmful sample-related corruption and improve the robustness of the ISTD model respectively. We also presented a flexible spatial-frequency interaction module to disentangle the salient features from the corrupted inputs. Extensive experiments  both on the general and degraded benchmarks demonstrate the superiority of our scheme with strong generalization ability. 

\bibliography{aaai24}

\end{document}